%% file: main.tex
\documentclass{article}

\usepackage{microtype}
\usepackage{graphicx}
\usepackage{subfigure}
\usepackage{booktabs} 
\usepackage{amsmath}
\usepackage{amssymb}
\usepackage{algorithm}
\usepackage{graphicx}
\usepackage{booktabs}
\usepackage{multirow}
\usepackage{array}
\usepackage{xcolor}
\usepackage{float}

\usepackage{amsmath}
\usepackage{amssymb}
\usepackage{amsthm}

\newtheorem{fplemma}{Lemma}
\newtheorem{fptheorem}{Theorem}
\usepackage{hyperref}

\usepackage[accepted]{mlsys2025}
\makeatletter
\def\Notice@String{}
\makeatother
\mlsystitlerunning{InfuseLoop: Cross-Trajectory Latent Injection for Scalable Parallel Looped Transformers}

\begin{document}

\twocolumn[
\mlsystitle{InfuseLoop: Cross-Trajectory Latent Injection for Scalable Parallel Looped Transformers}



\mlsyssetsymbol{Intern}{*}

\begin{mlsysauthorlist}
\mlsysauthor{Zhaofeng Sun}{THU,Intern}
\mlsysauthor{Lanxiang Hu}{UCSD}
\mlsysauthor{Yu-Yang Qian}{NJU}
\mlsysauthor{Jiaqi Leng}{UCSD}
\mlsysauthor{Yuanbo Yang}{UCSD}
\mlsysauthor{Chi Zhang}{THU}
\mlsysauthor{Youyou Lu}{THU}
\mlsysauthor{Peng Zhao}{NJU}
\mlsysauthor{Hao Zhang}{UCSD}
\end{mlsysauthorlist}

\mlsysaffiliation{THU}{Tsinghua University}
\mlsysaffiliation{UCSD}{University of California, San Diego}
\mlsysaffiliation{NJU}{Nanjing University}

\mlsyscorrespondingauthor{Hao Zhang}{haz094@ucsd.edu}

\mlsyskeywords{Looped Transformer, Efficient Architecture, Machine Learning Systems, Parallel Decoding}

\vskip 0.3in

\begin{abstract}
Parallel Looped Transformers (PLTs) use otherwise idle decoding compute by overlapping recurrent iterations. However, vanilla PLTs scale poorly with depth: they exhibit training-loss spikes, and their quality gains shrink as more loops are added. The benefit of additional loops depends on the information available to each iteration, so we examine whether the existing computational dependencies limit effective depth scaling. We find that only the preceding in-trajectory state is passed across loop boundaries, even though completed cross-trajectory representations are already available. Based on this finding, we propose InfuseLoop, a cross-trajectory latent injection mechanism. It aggregates causally available hidden states from later prediction trajectories through lightweight attention residuals. This design preserves token-parallel training and concurrent loop execution, and requires only $O(Rd)$ additional hidden-state storage. Experiments show that InfuseLoop eliminates training-loss spikes and improves validation-loss scaling up to eight loops. At eight loops, InfuseLoop reduces perplexity by 3.4--17.1\% relative to vanilla PLTs and by 10.5--37.3\% relative to the non-looped baseline. It also improves average downstream accuracy by 0.84 and 1.99 percentage points over these two baselines. InfuseLoop incurs only 0.3\% decoding time over vanilla PLTs and 23.3\% over the non-looped baseline, while keeping the same logical KV-cache footprint as vanilla PLTs.
\end{abstract}
]



\printAffiliationsAndNotice{%
    \textsuperscript{*}Work done during an internship
    at UC San Diego.\space
}  

\input{main/01_introduction}
\input{main/02_background}
\input{main/03_methodology}
\input{main/05_experiments}
\input{main/06_relatedwork}
\input{main/07_limitations}
\input{main/08_conclusions}

\bibliography{ref}
\bibliographystyle{mlsys2025}

\section*{AI Usage Disclosure}

The authors used AI-assisted tools, including Codex and Claude Code, to improve productivity during research and manuscript preparation. These tools primarily assisted with code implementation, debugging, data processing, experimental automation, visualization, and language refinement under the authors' supervision and verification. All research ideas, methodological innovations, experimental designs, scientific analyses, and conclusions were developed and determined by the authors. AI tools served solely as implementation and productivity aids. The authors reviewed and validated AI-assisted outputs and take full responsibility for the accuracy and integrity of this work.

\clearpage
\appendix
\input{main/appendix_A}
\input{main/appendix_B}

\end{document}

%% file: main/01_introduction.tex
\section{Introduction}

\begin{figure*}[!t]
    \centering
    \includegraphics[width=0.95\textwidth]
        {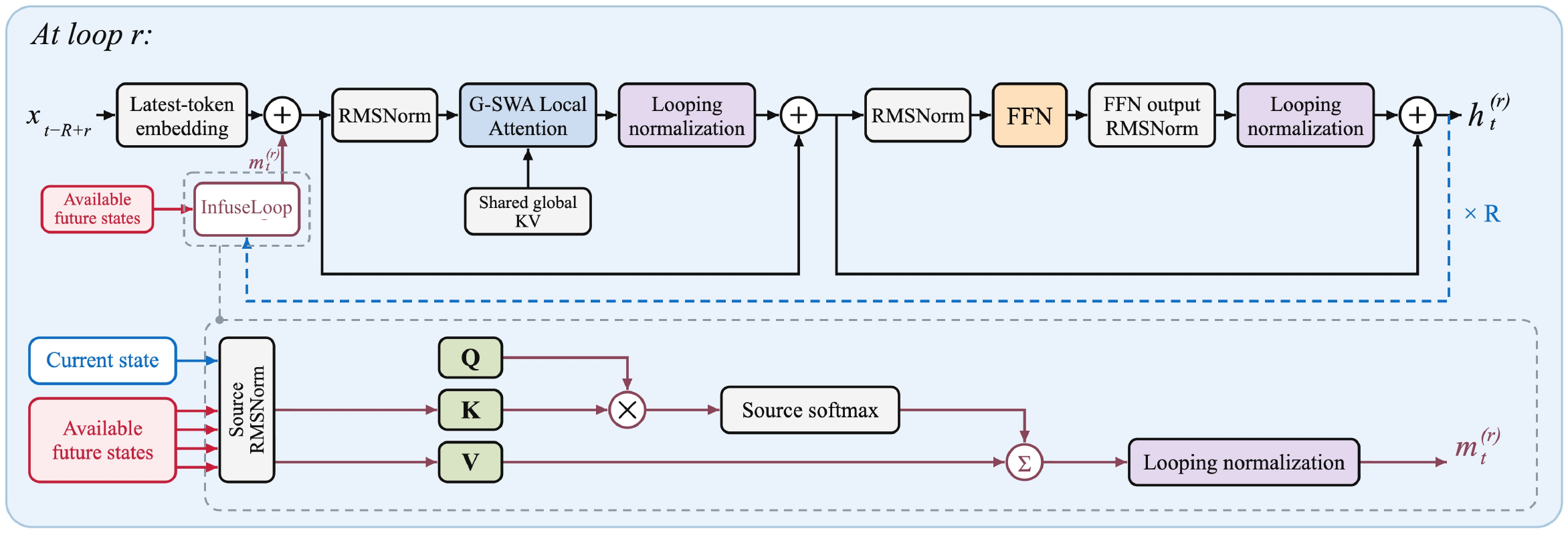}
    \caption{
        \textbf{Overview of InfuseLoop.}
        Each loop combines the latest token embedding with
        the current state and causally available future states
        through attention residual aggregation.
        The enlarged view details the attention-based aggregation of current latent state and future states.
    }
    \label{fig:foresight-overview}
\end{figure*}

Scaling computational depth can improve model quality, but adding layers to a standard Transformer also increases its parameter count, raising the memory and bandwidth costs of deployment~\cite{transformers}. This motivates using additional computation to improve model quality without increasing model size. Looped Transformers (LTs) achieve this by repeatedly applying shared Transformer blocks, so that effective depth grows while the parameter count stays fixed~\cite{ouro,huginn}. However, because LTs execute loops sequentially, they incur substantial inference overhead: both decoding latency and KV-cache memory grow roughly linearly with the number of loop iterations. This overhead limits how far recurrent depth can be scaled in practice.

Parallel Looped Transformers (PLTs) address this bottleneck by restructuring the dependencies across token positions and loop iterations, which allows different loop stages from multiple prediction trajectories to run concurrently~\cite{plt}. Combined with cross-loop KV-cache sharing, this parallel schedule keeps decoding latency and cache storage close to those of a non-looped model when decoding is memory-bound. PLTs therefore make it possible to improve model quality by using decoding compute that would otherwise be idle.

However, efficient execution alone does not guarantee effective depth scaling in PLTs. Prior work~\cite{loopcoder-v2} reports that performance degrades beyond two loops and attributes this to diminishing refinement gains. Our experiments reveal a further problem: vanilla PLTs show training-loss spikes with as few as four loops. Examining the computational graph, we find a gap between the states that are available and the states that the recurrent connections actually use. At each loop boundary, a vanilla PLT passes only the previous output of the same prediction trajectory to the next loop. Yet its overlapping execution schedule also produces completed states on other trajectories, and these states are causally available to the receiving loop. They provide additional representations of the observed context, but the loop-to-loop connections give no direct access to them. We therefore ask: can aggregating these available cross-trajectory representations make additional parallel computation more effective?

\begin{figure*}[t]
    \centering
    \begin{minipage}[t]{0.48\textwidth}
        \centering
        \includegraphics[width=\linewidth]
            {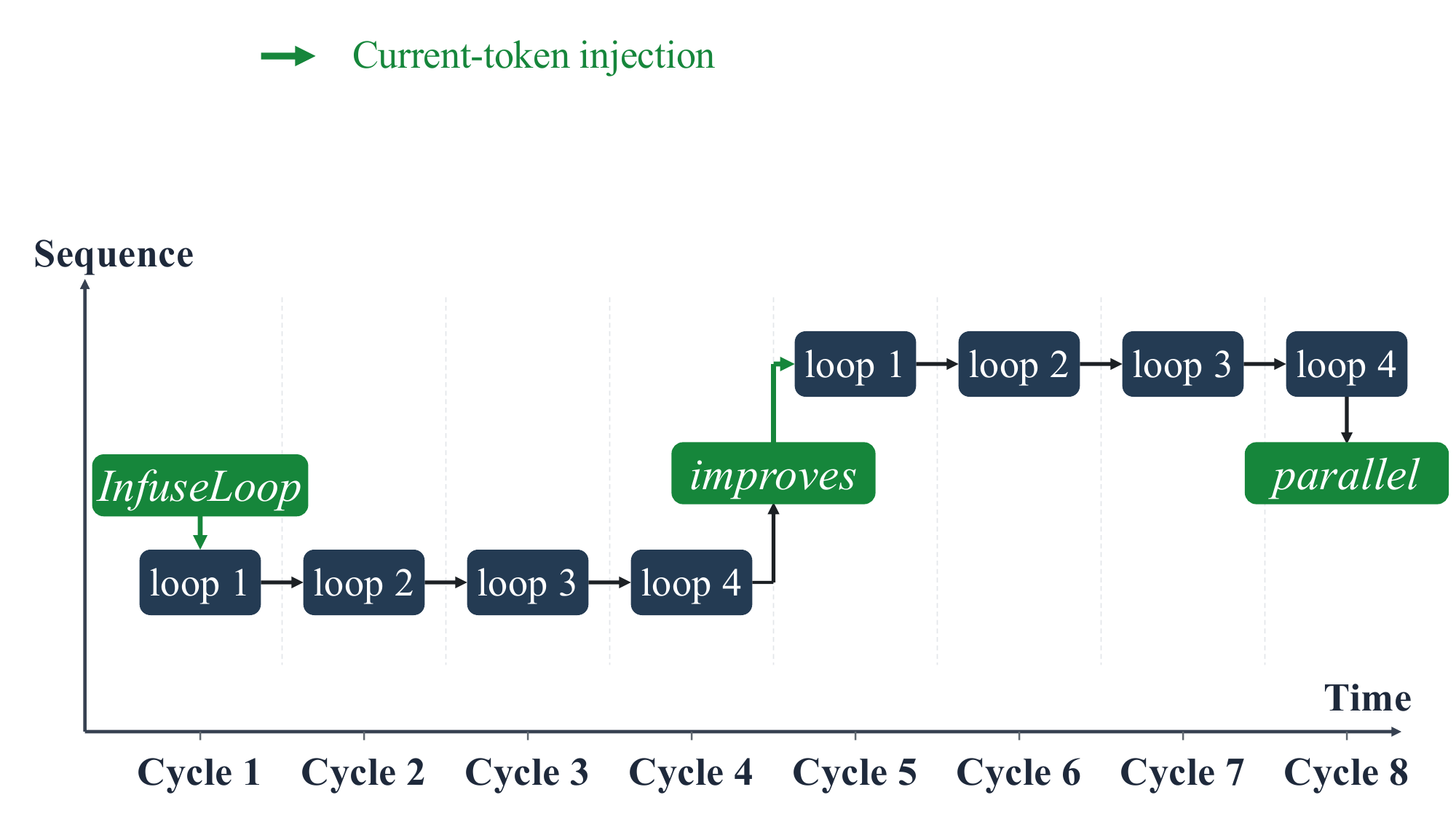}
        \par\vspace{2pt}
        {\small (a) Looped Transformer}
    \end{minipage}
    \hfill
    \begin{minipage}[t]{0.48\textwidth}
        \centering
        \includegraphics[width=\linewidth]
            {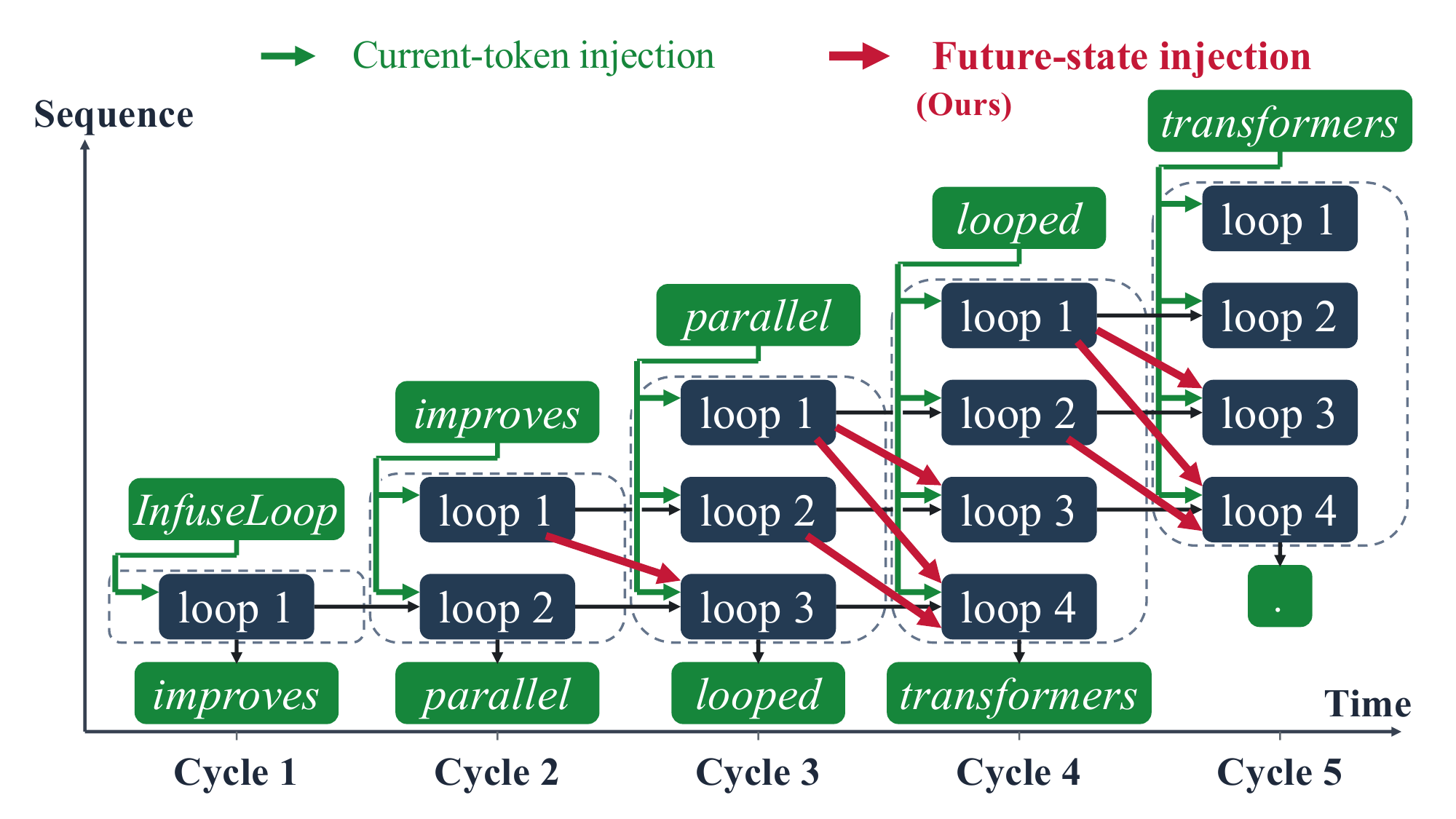}
        \par\vspace{2pt}
        {\small (b) InfuseLoop (Ours)}
    \end{minipage}

    \caption{
        \textbf{Execution schedules and information flow.}
        (a) Looped Transformers execute successive loops serially
        before generating the next token.
        (b) InfuseLoop overlaps computation across prediction
        positions and loop iterations, and introduces future-state
        injection (red) on top of vanilla PLT.
    }
    \label{fig:overview}
\end{figure*}

In this work, we introduce \textbf{InfuseLoop}, which is based on a simple intuition: earlier loops can perform more useful computation if they can directly combine more of the contextual representations that are already available. As illustrated in Figure~\ref{fig:overview}, InfuseLoop injects hidden states from shallower loops of later prediction trajectories into deeper loops of earlier trajectories. These additional connections allow earlier loop stages to use representations beyond their own recurrent state, so that the computation before the final prediction is used more effectively. Because every injected state is computed before it is used, InfuseLoop expands information flow across trajectories while preserving the parallel execution structure of PLTs.

To the best of our knowledge, InfuseLoop is the first PLT architecture to train stably and continue improving in quality up to eight loops. At eight loops, our model achieves lower validation loss than vanilla PLTs. Compared with vanilla PLTs and a non-looped baseline, it reduces perplexity by 3.4--17.1\% and 10.5--37.3\% and improves average downstream accuracy by 0.8 and 2.0 percentage points, respectively. InfuseLoop achieves these gains with less than 1\% additional decoding time over vanilla PLTs while keeping the same logical KV-cache footprint. These results show that cross-trajectory information flow is an effective way to improve depth scaling in PLTs. It allows additional recurrent computation to improve quality while preserving efficient parallel decoding across loops.

In summary, our contributions are as follows:
\begin{itemize}
    \item \textbf{Computational graph analysis of PLTs.}
    We find that, at each loop boundary, a vanilla PLT passes
    only the preceding in-trajectory state to the next loop,
    so completed cross-trajectory states are never used as
    direct recurrent inputs.
    We further show that selected shallower-loop states from
    the previous cycle can be incorporated into deeper-loop
    updates without breaking autoregressive causality,
    token-parallel training, or concurrent loop execution.

    \item \textbf{InfuseLoop: cross-trajectory aggregation.}
    We propose InfuseLoop, which aggregates causally
    available hidden states from future positions through
    lightweight attention residuals.
    InfuseLoop reduces training-loss spikes and improves
    model quality while preserving token-parallel training
    and concurrent loop execution, and it requires only
    $O(Rd)$ additional hidden-state storage.

    \item \textbf{Depth-scaling and system-level evaluation.}
    We evaluate depth scaling up to eight loops and
    characterize the quality--latency trade-offs, execution
    costs, and cache-memory overhead of InfuseLoop.
    The results show that InfuseLoop scales recurrent
    computation while remaining efficient at the system level.
\end{itemize}

%% file: main/02_background.tex
\section{Preliminary}
\label{sec:2}
\subsection{Looped Transformers Depth Scaling}

Looped Transformers (LTs)~\cite{universaltransformers}
decouple computational depth from parameter count by repeatedly
applying shared Transformer blocks.
Let $V$ be the vocabulary and $x_t \in V$ the input token at
position $t$.
Consider a standard full-stack LT with $K$ Transformer blocks
$f_{\theta_1},\ldots,f_{\theta_K}$, whose composition is
$F_{\theta}=f_{\theta_K}\circ\cdots\circ f_{\theta_1}$.
Starting from $h_t^{(0)}=\operatorname{Embed}(x_t)$, the model
applies this backbone for $R$ loops:
\begin{equation}
\begin{aligned}
    h_t^{(r)} &= F_{\theta}\!\left(h_t^{(r-1)}\right),
        \qquad r=1,\ldots,R, \\
    x_{t+1} &= \operatorname{LM\_Head}\!\left(h_t^{(R)}\right).
\end{aligned}
\end{equation}
Because the parameters $\theta=\{\theta_1,\ldots,\theta_K\}$
are shared across loops, the model performs $K\times R$
Transformer-block applications with only $K$ distinct blocks.

Empirical studies show that recurrent computation can improve model quality across architectures and training stages, from pretrained dense and mixture-of-experts (MoE) language models~\cite{huginn,smelt} to instruction-tuned models~\cite{ouro}. Realizing these gains, however, requires addressing the optimization and stability problems caused by reusing parameters across loops. Existing approaches address these problems with loop-aware residual scaling and initialization~\cite{deeploop,loop-mup}, constrained input injection~\cite{parcae}, and variable-depth training with intermediate supervision~\cite{huginn,ouro}. These advances make recurrent computation more effective, but its sequential execution still increases autoregressive decoding latency, which motivates parallel looping schemes.

\subsection{Parallel Looped Transformers}

Despite these benefits, LTs are difficult to deploy because each prediction requires $R$ sequential evaluations of the shared backbone $F_\theta$. As a result, autoregressive decoding latency grows roughly linearly with the number of loops, which limits the practical benefit of greater computational depth.

One way to reduce this sequential cost is to overlap the loop stages of different predictions. This overlap is possible only if each loop receives its recurrent input without waiting for another loop that runs at the same timestep. Parallel Looped Transformers (PLTs) enable this execution pattern by restructuring the dependencies across prediction trajectories~\cite{plt}. As illustrated in Figure~\ref{fig:overview}, a trajectory starts its computation from earlier context and incorporates newly available tokens as it moves through the loops. As a result, different trajectories can run different loop stages at the same time.

Consider a trajectory whose final loop runs at timestep $t$
and predicts $x_{t+1}$.
Note that $t$ indexes the last timestep of the trajectory,
not its first.
The trajectory computes
\begin{equation}
\label{eq:plt-computation}
\begin{aligned}
    h_t^{(1)} &= F_{\theta}
        \!\left(\operatorname{Embed}(x_{t-R+1})\right), \\
    h_t^{(r)} &= F_{\theta}
        \!\left(h_t^{(r-1)}
        + \operatorname{Embed}(x_{t-R+r})\right),
        \quad r=2,\ldots,R, \\
    x_{t+1} &= \operatorname{LM\_Head}
        \!\left(h_t^{(R)}\right).
\end{aligned}
\end{equation}
Loop $r$ runs at timestep $t-R+r$ and incorporates the token
available at that step, and the final state $h_t^{(R)}$
predicts $x_{t+1}$.
Therefore, unlike in a standard LT, a trajectory starts
computing before $x_t$, the token immediately preceding its
prediction target, is available.
This allows early loops of later trajectories to run at the
same time as deeper loops of earlier trajectories.
For example, the following states can be computed in
parallel at timestep $t$:
\begin{equation*}
\begin{aligned}
    h_{t+R-1}^{(1)} &= F_{\theta}
        \!\left(\operatorname{Embed}(x_t)\right), \\
    h_{t+R-2}^{(2)} &= F_{\theta}
        \!\left(h_{t+R-2}^{(1)} + \operatorname{Embed}(x_t)\right), \\
    &\;\;\vdots \\
    h_t^{(R)} &= F_{\theta}
        \!\left(h_t^{(R-1)} + \operatorname{Embed}(x_t)\right),
\end{aligned}
\end{equation*}
because each depends only on $x_t$ and on states computed at
earlier timesteps.

Although LTs and PLTs share weights across loops, their hidden representations differ from loop to loop, so standard implementations keep a separate KV cache for each loop. This increases the per-token cache footprint by a factor of $R$, which can increase memory traffic and limit serving batch sizes. KV-cache sharing can reduce this overhead. For example, gated sliding-window attention (G-SWA)~\cite{plt} shares the global KV cache of the first loop with all later loops, while each loop keeps its own bounded local window. Beyond looped models, YOCO~\cite{yoco} builds global KV representations in a self-decoder and reuses them across cross-decoder layers. These designs show that cache reuse can limit the memory cost of repeated attention, which complements methods that reduce the cost of sequential execution.

However, efficient execution does not guarantee effective depth scaling. Prior work reports diminishing refinement gains and even performance degradation beyond two loops~\cite{loopcoder-v2}. We therefore focus on enabling PLTs to benefit from more than two loops ($R > 2$).

\subsection{In-trajectory Aggregation and Cross-trajectory Aggregation}

Residual connections give later computation direct access
to intermediate representations~\cite{resnet}.
In recurrent Transformers, this kind of routing can be viewed
along two complementary axes:
\emph{in-trajectory aggregation}, which combines
representations across layers or loop iterations
within the same prediction trajectory%
~\cite{attnres,loopingbeyondtwice,infiloop}; and
\emph{cross-trajectory aggregation}, which passes
hidden representations between prediction positions%
~\cite{fullbandwidthtransformer,t2mlr}.

PLTs bring these two axes together through their overlapping execution schedule. With $R$ loops, each decoding cycle produces up to $R$ latent states, each at a different prediction position and recurrent depth. These states are different transformations of the observed context, not additional unobserved input tokens. Although attention already passes information across token positions, the explicit loop-output connections in a vanilla PLT carry only the previous state of the same trajectory. An important design question is therefore how to use these simultaneously available states without introducing dependencies that would serialize loop execution. In this work, we study how directly aggregating completed loop outputs across trajectories affects model quality and depth scaling, while preserving the execution schedule of PLTs.

%% file: main/03_methodology.tex
\section{InfuseLoop}

We propose InfuseLoop, which gives recurrent computation direct access to completed representations from overlapping prediction trajectories. Section~\ref{sec:foresight-graph} analyzes which states are available and describes how we select cross-trajectory sources. Section~\ref{sec:foresight-architecture} presents the aggregation architecture. Sections~\ref{sec:foresight-training} and~\ref{sec:foresight-inference} describe training and inference, and explain how the same design supports token-parallel training, concurrent loop execution, and bounded additional state storage.

\subsection{InfuseLoop Computational Graph}
\label{sec:foresight-graph}

Equation~\eqref{eq:plt-computation} describes a trajectory
that predicts $x_{t+1}$ through $R$ evaluations of the backbone.
The first $R-1$ evaluations run before $x_t$ is available,
so $x_t$ is injected only in the final loop.
At each loop transition, a vanilla PLT passes only the
previous output of the same trajectory to the next loop.
Its overlapping execution schedule also produces completed
states on other trajectories, but the recurrent update does
not take them as inputs.
This raises a natural question: can access to these
additional representations improve recurrent computation
without changing the parallel execution schedule?

InfuseLoop adds direct connections from later prediction
trajectories to earlier ones, using only states that have
already been computed.
For example, $h_{t+1}^{(1)}$ is computed at cycle $t-R+2$,
one cycle before $h_t^{(3)}$, so it can be aggregated with
$h_t^{(2)}$ to form the input to the third loop.
Although $h_{t+1}^{(1)}$ belongs to a later trajectory, it
encodes only the observed context and has no access to
$x_{t+1}$.
More generally, these connections do not make $x_t$ available
before cycle $t$; they only enlarge the set of contextual
representations that recurrent updates can use before the
final loop conditions on $x_t$.
Our experiments evaluate how this change affects optimization
behavior and predictive quality.

\begin{figure}[htbp]
    \centering
    \includegraphics[width=\linewidth]
        {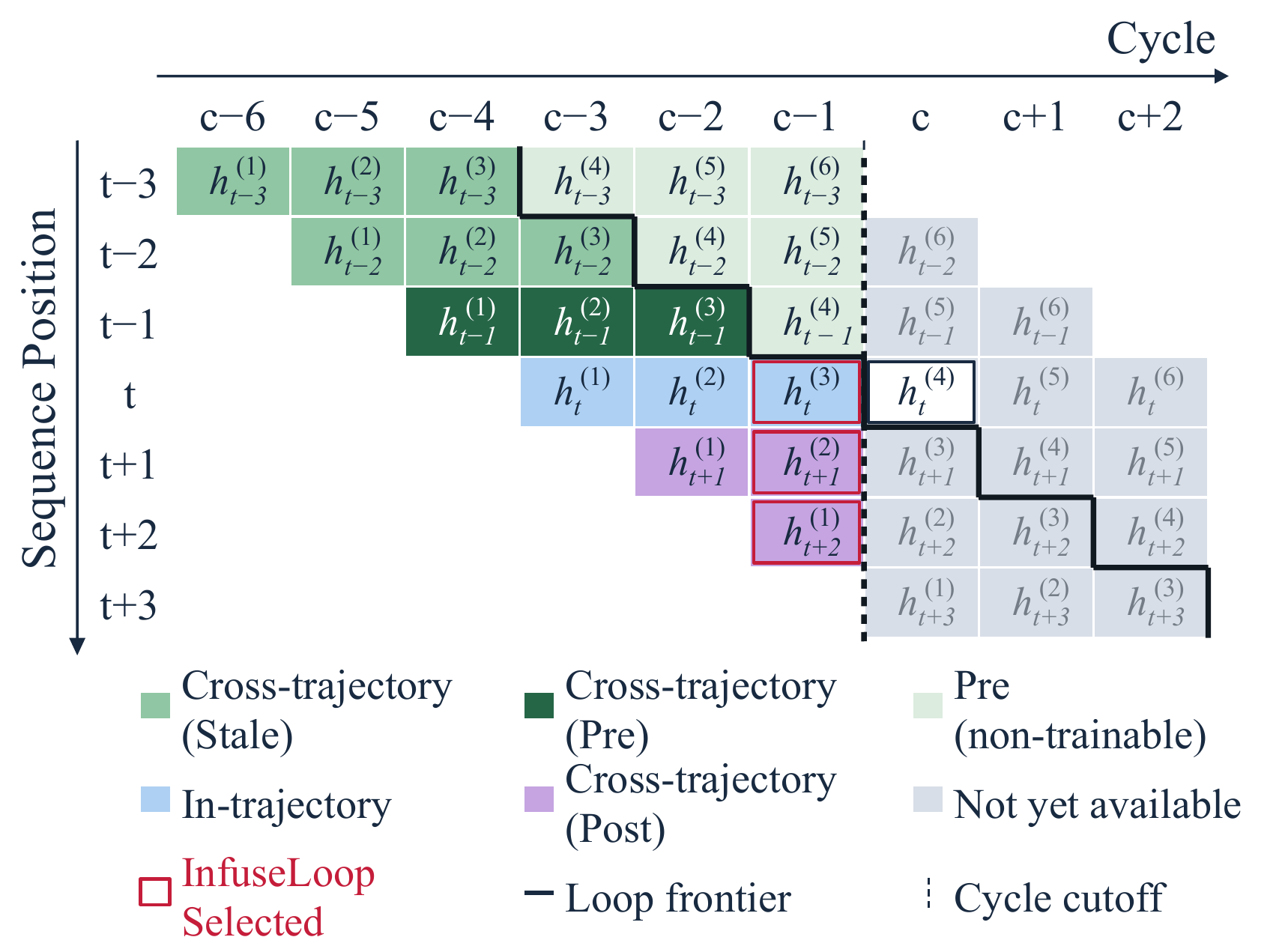}
    \caption{
        \textbf{State availability and source selection} for
        $h_t^{(4)}$ in a six-loop model ($R=6$).
        $h_t^{(4)}$ is computed at cycle $c=\tau(t,4)$, and the
        outlined states are the sources selected from cycle $c-1$.
    }
    \label{fig:plt-computational-graph}
\end{figure}

Figure~\ref{fig:plt-computational-graph} illustrates which
states are available when computing $h_t^{(4)}$.
Following Section~\ref{sec:2}, let $\tau(t,r)=t-R+r$ denote
the execution cycle of $h_t^{(r)}$.
We impose two constraints on an aggregation source $h_s^{(j)}$.
First, $\tau(s,j)<\tau(t,r)$: the source must be completed
in an earlier cycle, so that no new dependencies arise
between loops running in the same cycle.
Second, $j<r$: the source must come from a shallower loop,
so that it is available under loop-ordered training, in
which all sequence positions of a loop are processed in
parallel.
Together, these constraints define the admissible source set
\begingroup
\setlength{\abovedisplayskip}{2pt}
\setlength{\belowdisplayskip}{2pt}
\setlength{\abovedisplayshortskip}{1pt}
\setlength{\belowdisplayshortskip}{1pt}
\begin{equation}
\label{eq:admissible-sources}
    \mathcal{A}_{t,r}
    =\left\{
        h_s^{(j)}:
        1\le j<r,\ s+j<t+r
      \right\}.
\end{equation}
\endgroup
Relative to the receiving trajectory $t$, admissible sources
fall into three groups: In-trajectory States from the same
trajectory ($s=t$), Pre States from earlier trajectories
($s<t$), and Post States from later trajectories ($s>t$).
These groups are defined by trajectory order, not by
execution time.
The figure also marks older Pre States (Stale) and Pre States with
$j\ge r$ (non-trainable).
The latter are causally available but are excluded, because they violate the loop
order required for token-parallel training.

Rather than aggregating the full admissible set, whose size
grows with sequence length, InfuseLoop selects all
shallower-loop outputs from the immediately preceding cycle:
\begingroup
\setlength{\abovedisplayskip}{2pt}
\setlength{\belowdisplayskip}{2pt}
\setlength{\abovedisplayshortskip}{1pt}
\setlength{\belowdisplayshortskip}{1pt}
\begin{equation}
\label{eq:foresight-sources}
\mathcal{S}_{t,r}
= \{h_{t+r-j-1}^{(j)} : 1 \le j < r\}.
\end{equation}
\endgroup

For example,
$\mathcal{S}_{t,4}=\{h_t^{(3)}, h_{t+1}^{(2)}, h_{t+2}^{(1)}\}$.
Each selected state satisfies
$\tau(t+r-j-1,j)=\tau(t,r)-1$ and $j<r$, and therefore
belongs to $\mathcal{A}_{t,r}$.
The source with $j=r-1$ is the in-trajectory base state
$h_t^{(r-1)}$, and the remaining $r-2$ sources are completed
outputs of later trajectories.
No Pre State is selected, because every admissible Pre State
was computed at least two cycles earlier.
Because all sources come from the previous cycle, a rolling
buffer of $O(R)$ hidden vectors suffices for all receiving
loops.
Appendix~\ref{app:future-state-routing} proves that the
selected states are causally available, that the selection
is compatible with the training and inference schedules,
and that the storage bound holds.

\subsection{Architecture Design}
\label{sec:foresight-architecture}

Inspired by Attention Residuals~\cite{attnres}, we aggregate
the selected sources with a learned pseudo-query
$w_r\in\mathbb{R}^{d}$ for each receiving loop $r$.
The keys and values are both obtained by applying
parameter-free RMS normalization to the source states:
\begingroup
\setlength{\abovedisplayskip}{2pt}
\setlength{\belowdisplayskip}{2pt}
\setlength{\abovedisplayshortskip}{1pt}
\setlength{\belowdisplayshortskip}{1pt}
\begin{equation}
\begin{aligned}
    k_{t,j\to r}
    &=v_{t,j\to r}
      =\operatorname{RMSNorm}\!\left(
          h_{t+r-j-1}^{(j)}
        \right),\\
    e_{t,j\to r}
    &=\frac{w_r^\top k_{t,j\to r}}{\sqrt{d}}+b_{r,j},\\
    \alpha_{t,j\to r}
    &=\frac{\exp(e_{t,j\to r})}
           {\sum_{i=1}^{r-1}\exp(e_{t,i\to r})},
\end{aligned}
\end{equation}
\endgroup
where $b_{r,j}$ is a learned bias for source $j$.
The aggregated residual $m_t^{(r)}$ then replaces
$h_t^{(r-1)}$ as the recurrent input:
\begingroup
\setlength{\abovedisplayskip}{2pt}
\setlength{\belowdisplayskip}{2pt}
\setlength{\abovedisplayshortskip}{1pt}
\setlength{\belowdisplayshortskip}{1pt}
\begin{equation}
\begin{aligned}
    m_t^{(r)}
    &=\rho_r\sum_{j=1}^{r-1}
           \alpha_{t,j\to r}v_{t,j\to r},\\
    h_t^{(r)}
    &=F_\theta\!\left(
          \operatorname{Embed}(x_{t-R+r})+m_t^{(r)}
        \right),
        \quad r=2,\ldots,R,
\end{aligned}
\end{equation}
\endgroup
where $\rho_r$ is a learned aggregation scale.
The first loop is unchanged.
At $r=2$, the only source is the in-trajectory state
$h_t^{(1)}$; Post States are included from $r=3$ onward.
Sources that cross a document boundary are masked before
the softmax, and an empty source set yields a zero residual.

We keep the G-SWA design of PLT~\cite{plt}.
In each backbone layer, the first loop provides global KV
representations that are shared with all later loops.
Each later loop combines attention over these shared
representations with its own sliding-window attention
through a learned gate.
For clarity, we omit the causal masks and KV-cache
arguments from $F_\theta$.

To stabilize recurrent computation, we apply loop-aware residual scaling to the attention and FFN updates~\cite{deeploop,loopingbeyondtwice}. Following the normalization design of looped MoE models~\cite{loopingbeyondtwice}, we also apply RMSNorm to the FFN output before the residual addition. Together with the RMS normalization of the sources and the learned aggregation scale $\rho_r$, these choices control the magnitudes of both the backbone updates and the cross-trajectory residuals. Section~\ref{sec:component-ablation} evaluates the contributions of these components separately from that of source selection.

\subsection{Training}
\label{sec:foresight-training}

\begin{algorithm}[!tbph]
\caption{Token-parallel training of InfuseLoop}
\label{alg:foresight-training}
\begin{algorithmic}[1]
\REQUIRE
\begin{tabular}[t]{@{}l@{}}
    Input sequence $X=(x_1,\ldots,x_L)$,\\
    next-token targets $Y$,\\
    number of loops $R$, backbone $F_\theta$
\end{tabular}
\STATE $\mathbf{E}\gets\operatorname{Embed}(X)$
    \COMMENT{Token embeddings}
\STATE $\mathbf{H}^{(1)}\gets F_\theta(\mathbf{E})$
    \COMMENT{First-loop states}
\FOR{$r=2,\ldots,R$}
    \FOR{$j=1,\ldots,r-1$}
        \STATE $\widetilde{\mathbf{H}}^{(j)}
            \gets\operatorname{ShiftRight}
            (\mathbf{H}^{(j)},1)$
            \COMMENT{Previous-cycle states}
    \ENDFOR
    \STATE $\mathbf{M}^{(r)}\gets
        \operatorname{Agg}_r
        (\{\widetilde{\mathbf{H}}^{(j)}\}_{j<r})$
        \COMMENT{Aggregate sources}
    \STATE $\mathbf{H}^{(r)}\gets
        F_\theta(\mathbf{E}+\mathbf{M}^{(r)})$
        \COMMENT{All positions in parallel}
\ENDFOR
\STATE $\mathcal{L}\gets\operatorname{CrossEntropy}
    (\operatorname{LM\_Head}(\mathbf{H}^{(R)}),Y)$
\STATE Backpropagate $\mathcal{L}$ and update parameters
\end{algorithmic}
\end{algorithm}

During training, loops are processed sequentially, while all
sequence positions within a loop are computed in parallel.
This order is valid because Equation~\eqref{eq:foresight-sources}
restricts every selected source to a shallower loop ($j<r$),
so all sources for loop $r$ are computed before loop $r$ begins.
In Algorithm~\ref{alg:foresight-training}, uppercase bold
symbols denote sequence-level tensors whose rows are ordered
by execution cycle.
Shifting these tensors right by one position therefore gives
the previous-cycle states required by
Equation~\eqref{eq:foresight-sources}.
As a result, training computes exactly the same computational
graph as inference, without approximating any of its
dependencies.

We apply the next-token objective to the outputs of the final
loop and backpropagate through all recurrent and aggregation
paths.
Appendix~\ref{app:gswa-training} describes the G-SWA masks and
how global and local KV representations are handled during
training.

\subsection{Efficient Inference}
\label{sec:foresight-inference}

Algorithm~\ref{alg:foresight-inference} describes one decoding
cycle, using lowercase symbols for individual states.
Each loop builds its input from the current token and the
cached outputs of shallower loops from the previous cycle,
so aggregation introduces no dependencies on outputs of the
current cycle.
The backbone follows the layer-synchronous G-SWA schedule of
PLT: at each layer, the states of all active loops are batched
together, and the global KV representations of the first loop
are shared with the other loops.
The output of the final loop predicts the next token, and the
cached hidden states are updated only after all outputs of the
current cycle have been computed.

Keeping one output per loop requires only $O(Rd)$ additional hidden-state storage per sequence, independent of sequence length, whereas keeping all outputs over $L$ cycles would require $O(RLd)$. This bound excludes model parameters, temporary buffers, and the KV cache, whose size is determined by the G-SWA design of PLT~\cite{plt}.

For a batch of $B$ requests with the same number of loops, each steady-state cycle passes $B\times R$ loop states to the shared backbone, which allows batched kernels and CUDA Graph execution. Section~\ref{sec:inference-efficiency} evaluates the decoding cost and cache footprint.

\begin{algorithm}[!tbph]
\caption{One decoding cycle of InfuseLoop}
\label{alg:foresight-inference}
\begin{algorithmic}[1]
\REQUIRE
\begin{tabular}[t]{@{}l@{}}
    Current token $x_c$,\\
    previous-cycle outputs $p_1,\ldots,p_R$,\\
    number of loops $R$, backbone $F_\theta$
\end{tabular}
\STATE $e_c\gets\operatorname{Embed}(x_c)$
\STATE $z_1\gets e_c$
    \COMMENT{First-loop input}
\FOR{$r=2,\ldots,R$}
    \STATE $m_r\gets
        \operatorname{Agg}_r(p_1,\ldots,p_{r-1})$
        \COMMENT{Aggregate sources}
    \STATE $z_r\gets e_c+m_r$
        \COMMENT{Loop input}
\ENDFOR
\STATE $u_{1:R}\gets
    \operatorname{ParallelForward}_\theta(z_{1:R})$
    \COMMENT{All loops in parallel}
\STATE Generate $x_{c+1}$ from
    $\operatorname{LM\_Head}(u_R)$
\STATE $p_{1:R}\gets u_{1:R}$
    \COMMENT{Cache sources for the next cycle}
\STATE \textbf{return} $x_{c+1}$ and $p_{1:R}$
\end{algorithmic}
\end{algorithm}

%% file: main/05_experiments.tex
\section{Experiments}

\subsection{Experimental Setup}

\noindent\textbf{Models and Data.}
We study models with about 0.6B parameters, each with a shared
backbone of 28 Transformer layers and a hidden size of
$d=1{,}024$.
We vary the number of loops while keeping the backbone fixed,
so additional loops add recurrent computation but no new
untied Transformer layers.
This setup isolates how each architecture uses the shared
backbone across repeated evaluations. Appendix~\ref{app:model-configuration} gives the full model 
configurations.

All models are trained on FineWeb-Edu~\cite{finewebedu} with
the same tokenizer and data splits, using 6B tokens for the
main experiments and 1B tokens for the component ablations.

We conduct all benchmark evaluations using
\texttt{lm-evaluation-harness}~\cite{lm_eval_harness}.
For language modeling, we report word perplexity on
WikiText-2~\cite{wikitext} and final-word perplexity on
LAMBADA~\cite{lambada}.
For downstream tasks, we report accuracy on
MMLU~\cite{mmlu}, HellaSwag~\cite{hellaswag},
ARC-Easy (ARC-E) and ARC-Challenge (ARC-C)~\cite{arc},
PIQA~\cite{piqa}, WinoGrande (WG)~\cite{winogrande},
and OpenBookQA (OBQA)~\cite{openbookqa}.
MMLU is evaluated in the five-shot setting, and all other
tasks in the zero-shot setting.
We use length-normalized accuracy (\texttt{acc\_norm}) for
HellaSwag, ARC-E, ARC-C, and OBQA, and standard accuracy
(\texttt{acc}) for MMLU, PIQA, and WG.
\textit{Avg.} is the unweighted mean of the seven downstream
accuracies and does not include perplexity.

\noindent\textbf{Training Settings and Implementation Details.}
We train all models on NVIDIA GB200 and H200 GPUs with a
sequence length of 1,024 and a global batch size of 128
sequences.
We use AdamW with a learning rate of $3\times10^{-4}$, and all
other optimization settings are shared across the compared
models.
Within each comparison, all models are trained on the same
number of tokens, so training compute grows with the number
of loops.
Unless ablated, all looped models use the looping normalization
described in Section~\ref{sec:foresight-architecture}.

\noindent\textbf{Baselines.}
We compare InfuseLoop with three baselines that use the same
backbone: Single, a standard LT, and a vanilla PLT.
Single applies the backbone once, LT runs its loops
sequentially, and PLT overlaps computation across prediction
trajectories.
All looped models use G-SWA with a window size of $w=64$
during decoding.
The component ablations change either the normalization or
the source selection, while keeping the rest of the
architecture and the training settings fixed.

\begin{figure*}[!t]
    \centering
    \begin{minipage}[t]{0.40\textwidth}
        \centering
        \includegraphics[width=\linewidth]
            {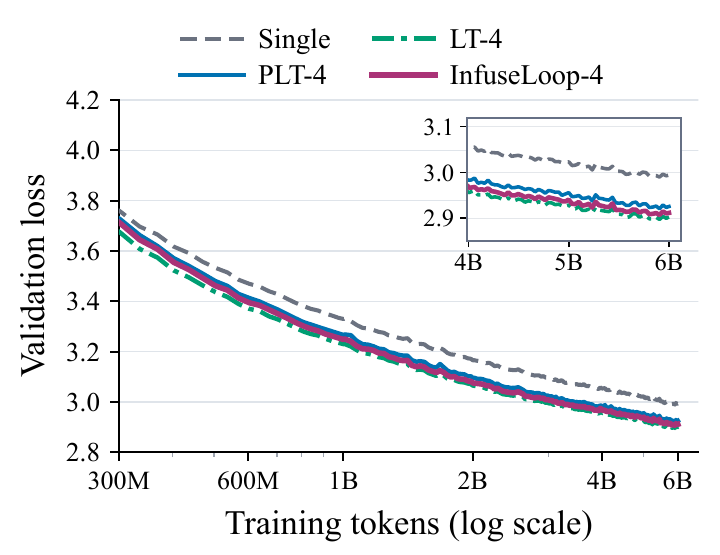}
        \par\vspace{4pt}
        {\small (a) Four loops}
    \end{minipage}
    \hspace{0.025\textwidth}%
    \begin{minipage}[t]{0.40\textwidth}
        \centering
        \includegraphics[width=\linewidth]
            {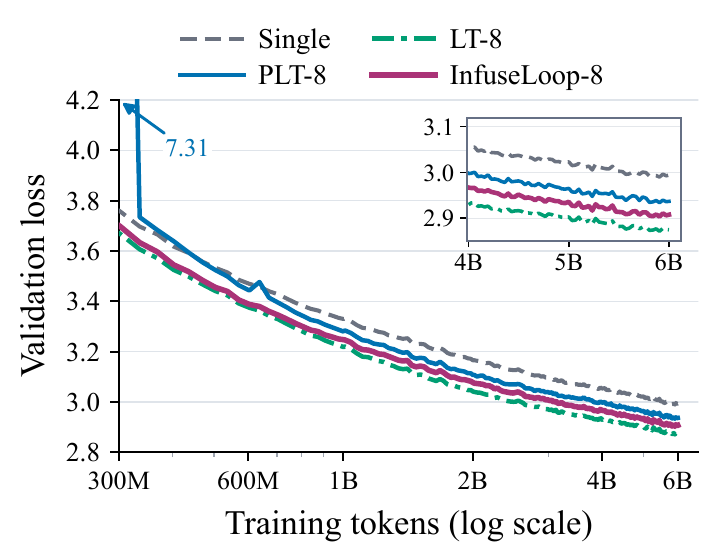}
        \par\vspace{4pt}
        {\small (b) Eight loops}
    \end{minipage}
    \caption{
        Validation loss during pretraining with
        (a) $R=4$ and (b) $R=8$ loops.
    }
    \label{fig:training-curves}
\end{figure*}

\subsection{Results}
\label{sec:results}

We compare the looped architectures at $R=4$ and $R=8$, using
Single as the non-looped reference.

Figure~\ref{fig:training-curves} shows validation loss during
pretraining.
Vanilla PLT shows pronounced loss spikes, whereas InfuseLoop
removes these spikes and converges more smoothly at both depths.
These results suggest that changing the restrictive
loop-to-loop information flow of vanilla PLT is an effective
way to stabilize parallel recurrent training.

Table~\ref{tab:main-results} reports validation loss, held-out
perplexity, and downstream accuracy.
At both depths, InfuseLoop outperforms vanilla PLT and Single
in validation loss, both perplexity metrics, and average
accuracy, and it closes \textbf{52--61\%} of the validation-loss
gap between vanilla PLT and LT.
Across both depths and both perplexity benchmarks, its
perplexity is \textbf{1.4--28.4\%} lower than that of vanilla PLT.
It also improves average downstream accuracy over vanilla PLT
by \textbf{0.96} percentage points at $R=4$ and by
\textbf{0.84} percentage points at $R=8$.
LT still achieves the highest average accuracy, but at eight
loops InfuseLoop trails it by only 0.59 percentage points,
recovering about \textbf{59\%} of the accuracy gap between
vanilla PLT and LT.
Thus, the benefits of cross-trajectory aggregation extend
beyond training stability to predictive quality.

\begin{table*}[t]
    \centering
    \caption{
        Validation loss, perplexity, and downstream accuracy (\%)
        after training on 6B tokens.
    }
    \label{tab:main-results}
    \small
    \setlength{\tabcolsep}{3pt}
    \begin{tabular}{@{}clc*{10}{c}@{}}
        \toprule
        \multirow{2}{*}{Loops}
            & \multirow{2}{*}{Model}
            & \multirow{2}{*}{Val. loss $\downarrow$}
            & \multicolumn{2}{c}{Perplexity $\downarrow$}
            & \multicolumn{8}{c}{Downstream tasks $\uparrow$} \\
        \cmidrule(lr){4-5}
        \cmidrule(lr){6-13}
        & &
            & Wiki2 & LAMBADA
            & MMLU & HellaSwag & ARC-E & ARC-C
            & PIQA & WG & OBQA & Avg. \\
        \midrule
        $R=1$ & Single
            & 2.9941
            & 39.72 & 67.02
            & 24.58 & 36.44 & 49.92 & 28.58
            & 63.38 & 51.70 & 33.00 & 41.09 \\
        \midrule
        \multirow{3}{*}{$R=4$}
            & LT
            & \underline{2.9022}
            & \underline{34.78} & \underline{41.07}
            & 24.35 & \underline{40.17}
            & \underline{54.34} & 30.55
            & \underline{66.43} & \underline{54.70}
            & \underline{33.80} & \underline{43.48} \\
            & PLT
            & 2.9252
            & 36.19 & 62.96
            & \underline{\textbf{24.71}} & 39.24
            & 51.68 & 28.58
            & 65.78 & 51.70 & 32.60 & 42.04 \\
            & \textbf{InfuseLoop}
            & \textbf{2.9112}
            & \textbf{35.70} & \textbf{45.08}
            & 24.49 & \textbf{39.49}
            & \textbf{53.75} & \underline{\textbf{31.40}}
            & \textbf{65.89} & \textbf{52.96}
            & \textbf{33.00} & \textbf{43.00} \\
        \midrule
        \multirow{3}{*}{$R=8$}
            & LT
            & \underline{2.8775}
            & \underline{33.78} & \underline{35.04}
            & 23.99 & \underline{41.88}
            & \underline{55.77} & \underline{31.91}
            & 65.51 & \underline{54.38}
            & 32.20 & \underline{43.66} \\
            & PLT
            & 2.9365
            & 36.82 & 50.69
            & 24.50 & 39.25 & 51.39 & 29.61
            & 65.72 & 52.33
            & \underline{\textbf{32.80}} & 42.23 \\
            & \textbf{InfuseLoop}
            & \textbf{2.9058}
            & \textbf{35.55} & \textbf{42.04}
            & \underline{\textbf{24.96}} & \textbf{40.00}
            & \textbf{53.20} & \textbf{31.43}
            & \underline{\textbf{66.16}} & \textbf{53.35}
            & 32.40 & \textbf{43.07} \\
        \bottomrule
    \end{tabular}

    \par\vspace{3pt}
    \begin{minipage}{\textwidth}
        \footnotesize
        \textit{Note:}
        Within each loop group, \underline{underlined} values are
        the best overall, and \textbf{bold} values are the best
        among non-LT models.
    \end{minipage}
\end{table*}

\subsection{InfuseLoop Efficiency and Scalability}
\label{sec:efficiency-scalability}

\subsubsection{Inference Efficiency}
\label{sec:inference-efficiency}

We benchmark decoding on a single NVIDIA GB200 GPU using
CUDA Graphs, the \texttt{vLLM} FA4 attention
backend~\cite{vllm}, and custom Triton kernels for G-SWA.
Linear and attention operands use FP16, while residuals,
reductions, and output logits use FP32.
After warmup and graph capture, we measure 64 consecutive
decoding cycles and report the median throughput over three
runs.
Component timings are collected separately from short CUDA
Graph execution traces.

\noindent\textbf{Memory footprint.}
Table~\ref{tab:cache-memory} reports the logical cache
footprint per cached token with $w=64$.
It includes the shared global KV cache, the loop-specific
local windows, and the $R-1$ latent vectors used as
aggregation sources.
KV entries are stored in 16-bit precision and latent vectors
in FP32.
The footprint is reported in decimal KB and excludes model
weights, temporary buffers, and allocator padding.
Because the global KV cache is shared, no loop needs its own
full-context cache, and the local windows and latent vectors
have a fixed size per sequence.
Their relative overhead therefore decreases as the context
grows, falling to \textbf{5.5\%} for InfuseLoop-8 at
8,192 tokens.
Because every term scales with the number of sequences, the
per-token footprint does not depend on batch size.

\begin{table}[t]
    \centering
    \caption{\textbf{Logical cache footprint} per token (KB)
        with window size $w=64$. Parentheses show the overhead
        relative to Single.}
    \label{tab:cache-memory}
    \small
    \setlength{\tabcolsep}{3pt}
    \begin{tabular}{@{}lrrr@{}}
        \toprule
        & \multicolumn{3}{c}{Context length} \\
        \cmidrule(lr){2-4}
        Model & 512 & 2,048 & 8,192 \\
        \midrule
        Single
            & 114.69 & 114.69 & 114.69 \\
        \addlinespace[3pt]
        \textbf{InfuseLoop-4}
            & \shortstack[r]{157.72\\
                {\scriptsize $(+37.5\%)$}}
            & \shortstack[r]{125.45\\
                {\scriptsize $(+9.4\%)$}}
            & \shortstack[r]{117.38\\
                {\scriptsize $(+2.3\%)$}} \\
        \addlinespace[3pt]
        \textbf{InfuseLoop-8}
            & \shortstack[r]{215.10\\
                {\scriptsize $(+87.5\%)$}}
            & \shortstack[r]{139.79\\
                {\scriptsize $(+21.9\%)$}}
            & \shortstack[r]{120.96\\
                {\scriptsize $(+5.5\%)$}} \\
        \bottomrule
    \end{tabular}
\end{table}

\noindent\textbf{Arithmetic intensity.}
We estimate arithmetic intensity as the FLOPs of the dominant
projection and attention operations divided by the distinct
bytes of packed weights and KV inputs, assuming that each
element is read once.
At $B=32$, $R=8$, and a 4,096-token context, InfuseLoop
has an estimated intensity of \textbf{26.8 FLOPs/byte}.
This is far below the GB200 ridge point of
\textbf{312.5 FLOPs/byte}, which is the 2.5~PFLOP/s dense FP16
peak divided by the 8~TB/s HBM bandwidth.
Decoding therefore remains memory-bound even with eight loops.
Parallel execution reuses the backbone weights across loops,
and G-SWA shares the global KV cache across loop queries.
Additional loops can therefore use idle compute without a
proportional increase in memory traffic, so latency grows
modestly rather than by a factor of $R$.

\begin{figure*}[!t]
    \centering
    \begin{minipage}[t]{0.35\textwidth}
        \centering
        \includegraphics[width=\linewidth]
            {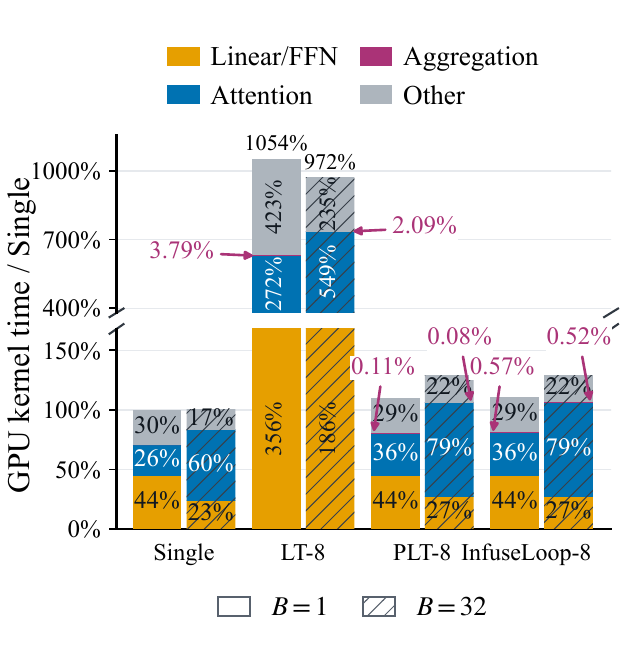}
        \par\vspace{3pt}
        {\small (a) Execution breakdown}
    \end{minipage}
    \hfill
    \begin{minipage}[t]{0.30\textwidth}
        \centering
        \includegraphics[width=\linewidth]
            {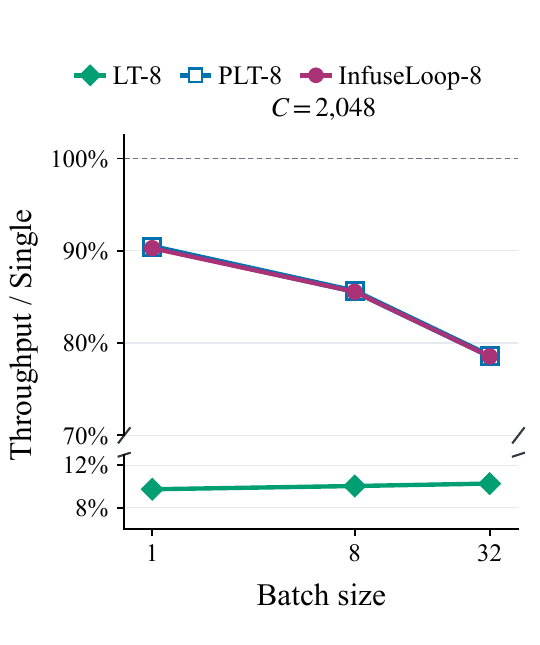}
        \par\vspace{3pt}
        {\small (b) Batch-size sensitivity}
    \end{minipage}
    \hfill
    \begin{minipage}[t]{0.30\textwidth}
        \centering
        \includegraphics[width=\linewidth]
            {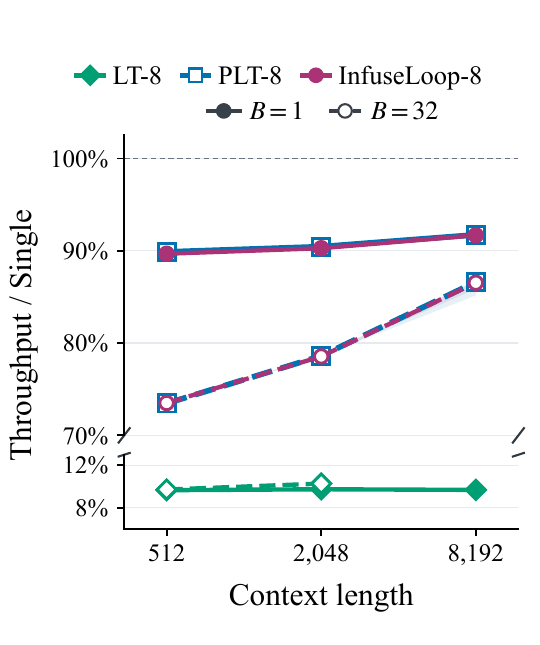}
        \par\vspace{3pt}
        {\small (c) Context-length sensitivity}
    \end{minipage}
    \caption{
        \textbf{Decoding cost and workload sensitivity.}
        (a) GPU kernel time with a 2,048-token initial context,
        normalized to the total kernel time of Single at the
        same batch size.
        (b, c) Throughput relative to Single across batch sizes
        and context lengths.
    }
    \label{fig:inference-analysis}
\end{figure*}

\noindent\textbf{Execution cost breakdown.}
Figure~\ref{fig:inference-analysis}(a) breaks down GPU kernel
time at $B=1$ and $B=32$ with a 2,048-token initial context,
normalized to the total kernel time of Single at the same
batch size.
Cross-trajectory aggregation accounts for less than
\textbf{0.6\%} of this reference time.
Because loops are batched, linear and FFN costs stay close to
those of Single, and most of the additional time comes from
attention.

\noindent\textbf{Batch and context sensitivity.}
Figure~\ref{fig:inference-analysis}(b) varies the batch size
with a 2,048-token initial context, and panel~(c) varies the
context length at $B=1$ and $B=32$.
The context grows during each measurement window, and all
throughput ratios are relative to Single at the same batch
size and context length.
InfuseLoop closely tracks vanilla PLT in all settings, as
expected from its small aggregation cost.
The throughput gap to Single widens with larger batches but
narrows with longer contexts.
At an 8,192-token context, InfuseLoop retains
\textbf{86.5\%} of the throughput of Single even at $B=32$.
This trend is consistent with the fixed costs of local windows
and aggregation becoming smaller relative to global attention
as the context grows.

\subsubsection{Depth Scalability}
\label{sec:depth-scalability}

Figure~\ref{fig:depth-pareto}(a) compares models trained
separately with $R=4$ and $R=8$, using the same backbone and
6B-token budget, with Single as the $R=1$ reference.
The horizontal axis is logarithmic, so successive doublings
of depth are equally spaced.
Because the data budget is fixed, this comparison measures
the quality gained from additional training compute.

Increasing vanilla PLT from four to eight loops worsens
validation loss, even though it doubles the number of
backbone applications.
In contrast, InfuseLoop reduces validation loss from four
to eight loops and outperforms vanilla PLT at both depths,
although its average downstream accuracy improves only
slightly.
LT still achieves the best quality, but InfuseLoop narrows
the gap without changing the parallel execution schedule.

\subsubsection{Performance--Latency Pareto Frontier}
\label{sec:loop-pareto}

\begin{figure*}[!t]
    \centering
    \begin{minipage}[t]{0.40\textwidth}
        \centering
        \includegraphics[width=\linewidth]
            {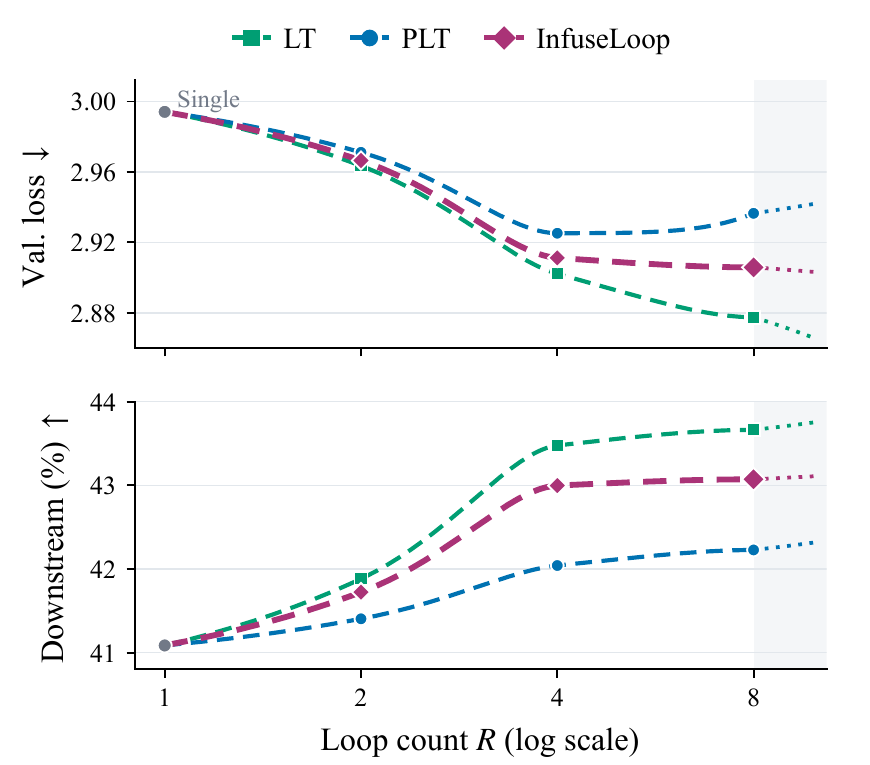}
        \par\vspace{3pt}
        {\small (a) Depth scaling}
    \end{minipage}
    \hspace{0.025\textwidth}%
    \begin{minipage}[t]{0.40\textwidth}
        \centering
        \includegraphics[width=\linewidth]
            {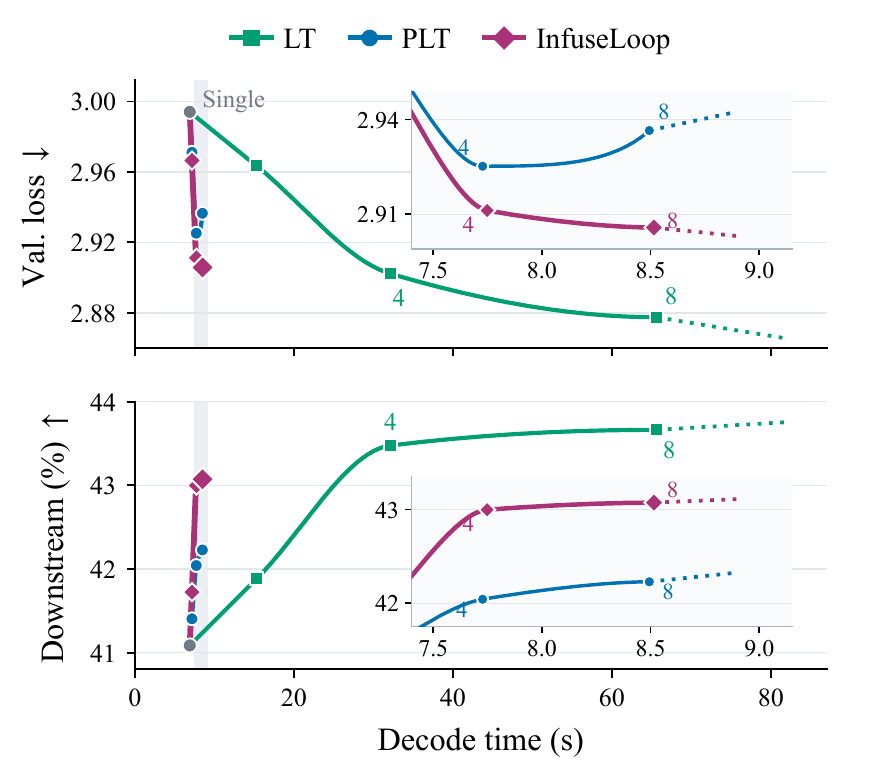}
        \par\vspace{3pt}
        {\small (b) Performance--latency Pareto frontier}
    \end{minipage}

    \caption{
        \textbf{Depth scaling and performance--latency Pareto
        frontiers.}
        Top: validation loss; bottom: average downstream
        accuracy.
        Interpolated $R=2$ points and projected $R>8$
        extensions are visual guides only; Pareto membership
        is computed from measured configurations.
    }
    \label{fig:depth-pareto}
\end{figure*}

Figure~\ref{fig:depth-pareto}(b) shows performance--latency
Pareto frontiers for validation loss and average downstream
accuracy.
We measure decode-only wall-clock time at $B=32$ with
2,048-token prompts and 2,048 output tokens per sequence,
timing the 2,047 cached steps after prefill.
We use the setup in Section~\ref{sec:inference-efficiency}
and report the median of three runs, excluding model loading,
warmup, and graph capture.
The loss and accuracy frontiers are computed separately from
measured configurations only.
A configuration is dominated if another is at least as good
and at least as fast, and strictly better in at least one of
the two.

InfuseLoop configurations lie on the measured Pareto frontier in the
low-latency regime.
InfuseLoop-4 dominates PLT-8: it achieves lower validation
loss and higher average accuracy at lower decoding latency.
InfuseLoop-8 further reduces validation loss at a modest
latency cost.
LT-8 achieves higher quality, but its decoding latency is
about \textbf{7.7$\times$} that of InfuseLoop-8.

LT remains nondominated in this fixed-backbone comparison,
but parameter efficiency does not imply latency
efficiency~\cite{plt}.
A larger non-looped model is an alternative use of the same
inference resources.
For example, at matched inference FLOPs, a 3.9B non-looped
model matches or exceeds a three-loop 1.4B model on all 12
benchmarks in a controlled comparison~\cite{closing_loop}.
InfuseLoop addresses this deployment cost: it obtains
quality gains from additional depth with a fixed backbone and
without the large latency increase of serial looping.


\subsection{Ablations}
\label{sec:ablations}
\label{sec:component-ablation}

Table~\ref{tab:component-ablation} evaluates source selection
and normalization at $R=4$ and $R=8$, using the same 0.6B
backbone trained on 1B tokens.

\begin{table}[t]
    \centering
    \caption{\textbf{Component ablations.}
        All configurations are trained on 1B tokens.}
    \label{tab:component-ablation}
    \small
    \setlength{\tabcolsep}{3pt}
    \renewcommand{\arraystretch}{1.08}
    \begin{tabular}{@{}p{0.42\linewidth}rrrr@{}}
        \toprule
        & \multicolumn{2}{c}{$R=4$}
        & \multicolumn{2}{c}{$R=8$} \\
        \cmidrule(lr){2-3}
        \cmidrule(lr){4-5}
        Configuration
            & Loss $\downarrow$ & Avg.\ $\uparrow$
            & Loss $\downarrow$ & Avg.\ $\uparrow$ \\
        \midrule
        \textbf{InfuseLoop}
            & 3.2482 & 38.26 & 3.2436 & 38.35 \\
        \midrule
        \multicolumn{5}{@{}l}{\textit{Source selection}} \\
        \quad Base only
            & 3.2660 & 38.04 & 3.2799 & 37.59 \\
        \quad FullState
            & 3.3975 & 37.43 & 3.4244 & 37.21 \\
        \quad In-trajectory
            & 3.2735 & 38.19 & 3.2839 & 37.87 \\
        \quad Pre
            & 3.2820 & 37.81 & 3.3707 & 37.41 \\
        \midrule
        \multicolumn{5}{@{}l}{\textit{Normalization}} \\
        \quad w/o source RMSNorm
            & 3.2883 & 38.02 & 3.3550 & 37.59 \\
        \quad w/o looping normalization
            & 3.2488 & 38.33 & 3.2434 & 38.17 \\
        \bottomrule
    \end{tabular}
\end{table}

\noindent\textbf{Source selection.}
Following the categories in Section~\ref{sec:foresight-graph},
we compare four alternative source sets: Base only, FullState,
In-trajectory, and Pre.
All variants keep the in-trajectory base state $h_t^{(r-1)}$
and use the same normalization and aggregation.
InfuseLoop achieves the lowest validation loss and the
highest average accuracy among these variants at both depths.
The Base-only and In-trajectory results show that information
from the same trajectory alone does not recover the benefit
of cross-trajectory aggregation.

FullState excludes stale Pre States but still aggregates
$O(R^2)$ candidates for deep receiving loops, compared with
$O(R)$ previous-cycle sources for InfuseLoop; all variants
use the same number of backbone evaluations.
Its higher loss suggests that a larger source set does not
necessarily improve quality at this training budget, possibly
because denser aggregation makes optimization harder.
Pre also underperforms, especially at $R=8$.
Its most recent eligible states are at least two cycles old,
compared with one cycle for InfuseLoop, which suggests that
staleness limits their usefulness.

\noindent\textbf{Normalization.}
Removing source RMSNorm worsens both validation loss and
average accuracy, especially at $R=8$.
Removing looping normalization, which consists of the
attention and FFN residual scaling and the FFN output RMSNorm,
while keeping source RMSNorm, leaves validation loss nearly
unchanged and changes average accuracy only slightly
($+0.07$ at $R=4$ and $-0.18$ at $R=8$).
Thus, at the 1B-token scale, source normalization affects
final quality more than the tested backbone normalization.

%% file: main/06_relatedwork.tex
\section{Related Work}

\subsection{Efficient Looped Transformers}

Efficient inference with looped models requires limiting both
repeated computation and the state stored for each loop.
Parameter-efficient designs combine layer tying with low-rank,
depth-specific adaptation~\cite{relaxed_recursive}, while
cache-compression methods use the structure shared across loops
through latent representations and quantized
residuals~\cite{looped_latent_attention,residualquant}.
Adaptive methods reduce redundant computation by choosing the
recursion depth for each token~\cite{mixture_of_recursions} or
by applying sparse updates across loops~\cite{flashloop}.
Other execution schemes increase parallelism through depth-wise
batching and pipelining across tokens~\cite{relaxed_recursive,plt},
or by predicting the inputs of later loops and correcting their
outputs afterward~\cite{loopfold}.
InfuseLoop keeps the pipelined schedule of PLT and improves
information flow across trajectories by aggregating completed
loop outputs.

The regularity of execution affects how well these designs map
to batched inference.
Adaptive depths can produce uneven workloads across requests,
which require dedicated batching and load-balancing
mechanisms~\cite{relaxed_recursive,mixture_of_recursions}.
In contrast, InfuseLoop keeps the regular execution pattern of PLT for requests with the same $R$, so each cycle has static shapes and supports CUDA Graph execution (Section~\ref{sec:foresight-inference}).

\subsection{Exploiting Memory-Bound Decoding}

When decoding is memory-bound, parallel computation can
amortize the cost of reading model weights and caches.
Speculative decoding drafts several tokens with a cheaper
model and verifies them in a single batched pass of the target
model~\cite{leviathan2023speculative,chen2023speculative}.
Auxiliary prediction heads and lightweight block-diffusion
drafters further reduce drafting cost and increase token-level
parallelism~\cite{medusa,dflash}.

InfuseLoop instead uses this spare compute for recurrent
computation that improves model quality, while keeping loop
execution parallel.
It is therefore potentially complementary to draft-and-verify
decoding.
However, because both approaches draw on the same spare
compute, their combined benefit depends on the workload and
the available compute.
For example, a lightweight block-diffusion drafter could
propose token blocks for a InfuseLoop target
model~\cite{dflash}.
Such a combination would require verification that accounts
for the loop pipeline, as well as consistent handling of
cached recurrent states when drafts are rejected.
We leave its implementation and end-to-end evaluation to
future work.

%% file: main/07_limitations.tex
\section{Limitations}

Because of our training compute budget, we evaluate only dense
models with about 0.6B parameters.
Scaling InfuseLoop to larger dense and mixture-of-experts
(MoE) models remains unexplored.

Although InfuseLoop improves depth scaling over vanilla PLT,
its quality gains from additional loops are still smaller than
those of sequential LTs at the evaluated depths.
Further study is needed to identify the architectural and
optimization factors behind this gap.

%% file: main/08_conclusions.tex
\section{Conclusions}

We presented InfuseLoop, an efficient architecture for
scaling parallel recurrent computation.
By injecting causally available hidden states from later
prediction trajectories, InfuseLoop improves training
stability and model quality while preserving token-parallel
training and concurrent loop execution.
Its lightweight aggregation and $O(Rd)$ hidden-state storage
add little decoding overhead over vanilla PLT.

Our results suggest that cross-trajectory aggregation is
important for deep recurrent models: additional computation
becomes more effective when representations from later
trajectories can reach the loops of earlier ones.
More broadly, InfuseLoop illustrates how redesigning these
information paths can improve depth scaling while retaining
the execution efficiency of parallel looping.

%% file: main/appendix_A.tex
\section{Causal Routing and Parallel Execution}
\label{app:future-state-routing}

This appendix establishes the causal availability,
execution compatibility, and auxiliary storage requirements
of InfuseLoop's source selection.
We use the notation of Section~\ref{sec:foresight-graph}:
$h_t^{(r)}$ is the output of loop $r$ on the trajectory
that predicts $x_{t+1}$, and its execution cycle is
$\tau(t,r)=t-R+r$.
The source characterization below assumes steady-state
execution.
Unavailable source states are masked, and an empty
source set produces a zero residual.

\subsection{Availability and Source Selection}
\label{app:routing-availability}

The admissible source set satisfies both the
execution-cycle and training-loop ordering constraints:
\begin{equation}
    \mathcal{A}_{t,r}
    =\left\{
        h_s^{(j)}:1\le j<r,\ s+j<t+r
      \right\}.
\end{equation}
InfuseLoop restricts this set to the immediately
preceding cycle:
\begin{equation}
    \mathcal{S}_{t,r}
    =\left\{
        h_{t+r-j-1}^{(j)}:1\le j<r
      \right\}.
\end{equation}

\begin{fplemma}[Previous-cycle source characterization]
\label{lem:previous-cycle-sources}
For a receiver $h_t^{(r)}$ with $r\ge2$,
$\mathcal{S}_{t,r}$ consists exactly of the admissible
sources produced at cycle $\tau(t,r)-1$.
It contains one in-trajectory state and $r-2$ Post states.
Any admissible Pre state is at least two cycles old.
\end{fplemma}

\begin{proof}
Let $c=\tau(t,r)$.
A state $h_s^{(j)}$ is produced at cycle $c-1$ if and
only if
\begin{equation}
    s-R+j=t-R+r-1,
\end{equation}
or equivalently $s=t+r-j-1$.
For every $1\le j<r$, this source belongs to
$\mathcal{A}_{t,r}$, since $s+j=t+r-1<t+r$.
Moreover, $s=t$ when $j=r-1$, while $s>t$ for
$1\le j\le r-2$.
Thus, the $r-1$ selected sources comprise the base
$h_t^{(r-1)}$ and $r-2$ completed Post states.

For any Pre source with $s<t$ and $j<r$,
\begin{equation}
    \tau(t,r)-\tau(s,j)
    =(t-s)+(r-j)\ge2.
\end{equation}
Hence no admissible Pre state was produced in the
immediately preceding cycle.
\end{proof}

This characterization explains the temporal distinction
between the selected Post sources and admissible Pre
sources; it does not determine their relative predictive
utility.
At $r=2$, selection contains only the in-trajectory base
state, and cross-trajectory aggregation begins at $r=3$.

\subsection{Autoregressive Causality}
\label{app:routing-causality}

\begin{fplemma}[Prefix causality]
\label{lem:prefix-causality}
For fixed model parameters, a state $h_t^{(r)}$ produced
at cycle $c=\tau(t,r)$ depends only on input tokens
observed by cycle $c$.
Future-state aggregation does not introduce dependence
on $x_{c+1},x_{c+2},\ldots$.
\end{fplemma}

\begin{proof}
We proceed by induction over execution cycles, and
within each cycle over Transformer layers.
Initially, unavailable source states are masked.
At cycle $c$, Lemma~\ref{lem:previous-cycle-sources}
shows that every aggregation source was produced at
cycle $c-1$.
By the induction hypothesis, these states depend only
on tokens already observed by that cycle.
Combining them with $\operatorname{Embed}(x_c)$ therefore
produces inputs depending only on the observed prefix
$x_{\le c}$.

At each backbone layer, causal G-SWA reads first-loop
global KV entries and loop-specific local entries only
from positions at or before $c$.
The current first-loop KV entry is computed from that
loop's layer input, which is itself causal by induction
over layers.
Attention, normalization, and FFN operations therefore
preserve dependence only on the observed prefix.
This establishes the claim for every current-cycle
loop output.
\end{proof}

Consequently, a larger trajectory index denotes a later
prediction, rather than access to a later input token.
In particular, for $R\ge3$, the connection
$h_{t+1}^{(1)}\rightarrow h_t^{(3)}$ reads a source
computed at $t-R+2$ before a receiver at $t-R+3$;
it does not provide $x_t$ before cycle $t$.

\subsection{Training and Inference Compatibility}
\label{app:routing-schedules}

\begin{fptheorem}[Equivalent evaluation orders]
\label{thm:routing-schedules}
Conditioned on the same input sequence and using the
same parameters, attention masks, and initial states,
loop-ordered token-parallel training and cycle-ordered
layer-synchronous inference evaluate the same InfuseLoop dependency graph.
The routing does not require an approximation to the
training dependencies or an additional serial loop stage
within a decoding cycle.
\end{fptheorem}

\begin{proof}
Consider a receiving state $h_t^{(r)}$ at cycle
$c=\tau(t,r)$.
By Lemma~\ref{lem:previous-cycle-sources}, each selected
source $h_{t+r-j-1}^{(j)}$ was produced at cycle $c-1$.
In Algorithm~\ref{alg:foresight-training}, sequence-level
tensors are ordered by execution cycle.
A one-position right shift therefore aligns the
loop-$j$ output from cycle $c-1$ with the receiving
position at cycle $c$.
Since $j<r$, the source tensor is already available
when training processes loop $r$.
Aggregation acts independently at each position,
and causal attention evaluates all sequence positions
within each backbone layer in parallel.

In Algorithm~\ref{alg:foresight-inference}, $p_j$ is
the cached output of loop $j$ from cycle $c-1$.
The states $p_1,\ldots,p_{r-1}$ thus supply exactly
the same sources as the shifted training tensors.
All current loop inputs can consequently be prepared
without reading current-cycle loop outputs.
The layer-synchronous backbone schedule then computes
the same layer operations: the first loop provides
each layer's global KV entries, and the remaining
loops use the same causal global and local attention
masks.
Current-layer projections may be batched before
attention; they do not require another loop to
complete its full backbone evaluation.

Thus, both schedules respect every aggregation and
backbone dependency and apply the same operations to
the same inputs.
Matching masks and initial states extend the argument
to positions with unavailable history.
The resulting states agree in exact arithmetic;
different kernel evaluation orders can introduce
ordinary floating-point differences.
\end{proof}

The statement compares evaluation on a fixed token prefix.
It does not remove the usual distinction between
teacher-forced training and autoregressive generation.
Training retains gradients through every selected
source, so token parallelism requires neither
stop-gradient operations nor truncated routing paths.

\subsection{Auxiliary State Storage}
\label{app:routing-storage}

At cycle $c$, define the retained source buffer as
\begin{equation}
    \mathcal{B}_{c-1}
    =\left\{
        h_{c+R-j-1}^{(j)}:1\le j\le R
      \right\}.
\end{equation}
Every receiving loop reads a subset of this buffer:
\begin{equation}
    \mathcal{S}_{c+R-r,r}
    =\left\{
        h_{c+R-j-1}^{(j)}:1\le j<r
      \right\}.
\end{equation}
The previous-cycle final-loop output is not read by
aggregation; retaining all $R$ outputs gives a uniform
cache update matching Algorithm~\ref{alg:foresight-inference}.
After the current cycle, the $R$ new loop outputs
replace the retained states.
For hidden dimension $d$ and batch size $B$, this
requires $O(BRd)$ persistent auxiliary storage,
independent of the generated sequence length.
Separate old and new buffers change only the constant
factor.
This bound excludes KV caches, model weights, training
activations, and temporary kernel workspaces.

There are $\sum_{r=2}^{R}(r-1)=R(R-1)/2$ logical
source--receiver pairs per sequence and cycle.
Their scoring and weighted reduction require
$O(R^2d)$ arithmetic, but no additional backbone
evaluations and no persistent $R\times R\times d$
source tensor.
This distinction separates aggregation work from
retained-state storage.

\paragraph{Scope of the structural results.}
These results establish how cross-trajectory aggregation
can be executed causally with the training and inference
orders used in PLT.
They do not imply an expressivity separation or a
guaranteed improvement in optimization or prediction.
The effects of source selection on training stability,
validation loss, and downstream performance are
assessed by the experiments and ablations.

%% file: main/appendix_B.tex
\section{Training Details}
\label{app:trainingdetails}

\subsection{Model Configuration}
\label{app:model-configuration}

All models follow the backbone configuration of
\texttt{Qwen3-0.6B} and use its tokenizer.
Table~\ref{tab:model-configuration} summarizes the
shared architectural parameters.
The input embedding and output projection share weights.
Looped variants reuse the same 28-layer backbone across
iterations, increasing computational depth without
replicating backbone parameters.
The total parameter count remains approximately
0.6B, with small additional parameter overhead
from architecture-specific aggregation modules.

\begin{table}[!hbp]
    \centering
    \caption{Backbone configuration and parameter counts.}
    \label{tab:model-configuration}
    \small
    \setlength{\tabcolsep}{4pt}
    \begin{tabular}{@{}lr@{}}
        \toprule
        Configuration & Value \\
        \midrule
        Backbone configuration & \texttt{Qwen3-0.6B} \\
        Transformer layers ($K$) & 28 \\
        Hidden dimension ($d$) & 1,024 \\
        FFN intermediate dimension & 3,072 \\
        Query heads & 16 \\
        KV heads & 8 \\
        Head dimension & 128 \\
        Vocabulary size & 151,936 \\
        FFN activation & SwiGLU \\
        Positional encoding & RoPE \\
        RoPE base & $10^6$ \\
        RMSNorm $\epsilon$ & $10^{-6}$ \\
        Tied input/output embeddings & Yes \\
        Training sequence length & 1,024 \\
        \midrule
        Base total parameters & 596,136,384 \\
        Embedding parameters & 155,582,464 \\
        Base non-embedding parameters & 440,553,920 \\
        \midrule
        InfuseLoop-4 total parameters & 596,139,465 \\
        InfuseLoop-8 total parameters & 596,143,587 \\
        \bottomrule
    \end{tabular}
    \par\vspace{3pt}
    \begin{minipage}{\linewidth}
        \footnotesize
        \textit{Note:} Base counts exclude cross-loop
        aggregation parameters. The tied embedding and
        output projection are counted once.
    \end{minipage}
\end{table}

\subsection{G-SWA During Training}
\label{app:gswa-training}

Following PLT~\cite{plt}, we use gated sliding-window
attention (G-SWA) to combine shared global context
with loop-specific local representations.
Training tensors are aligned by execution cycle:
row $c$ at loop $r$ corresponds to the computation
that produces $h_{c+R-r}^{(r)}$.
For backbone layer $\ell=1,\ldots,K$, let
$\mathbf{Q}_{\ell}^{(r)}$,
$\mathbf{K}_{\ell}^{(r)}$, and
$\mathbf{V}_{\ell}^{(r)}$ denote the sequence-level
query, key, and value tensors at loop $r$.
We suppress batch and attention-head indices below.

\noindent\textbf{Shared global and loop-specific local KV.}
The first loop performs causal self-attention and
constructs a separate global KV bank at each layer:
\begin{equation}
    \bigl(
        \mathbf{K}_{\ell}^{\mathrm{g}},
        \mathbf{V}_{\ell}^{\mathrm{g}}
    \bigr)
    =
    \bigl(
        \mathbf{K}_{\ell}^{(1)},
        \mathbf{V}_{\ell}^{(1)}
    \bigr).
\end{equation}
These tensors are constructed once per training
forward pass and reused by all subsequent loops
at the corresponding layer.
Each later loop computes its own queries and local
KV representations.
For $r>1$, the two attention branches are
\begin{equation}
\begin{aligned}
    \mathbf{A}_{\ell,\mathrm{g}}^{(r)}
    &=
    \operatorname{Attn}\!\left(
        \mathbf{Q}_{\ell}^{(r)},
        \mathbf{K}_{\ell}^{\mathrm{g}},
        \mathbf{V}_{\ell}^{\mathrm{g}};
        \mathbf{M}^{\mathrm{g}}
    \right), \\
    \mathbf{A}_{\ell,\mathrm{loc}}^{(r)}
    &=
    \operatorname{Attn}\!\left(
        \mathbf{Q}_{\ell}^{(r)},
        \mathbf{K}_{\ell}^{(r)},
        \mathbf{V}_{\ell}^{(r)};
        \mathbf{M}^{\mathrm{loc}}
    \right),
\end{aligned}
\label{eq:gswa-training-branches}
\end{equation}
where $\operatorname{Attn}$ denotes scaled
dot-product attention with an additive mask.
The two branches normalize their attention
distributions independently.

\noindent\textbf{Causal masking and gated combination.}
Let $D_c$ identify the document segment containing
position $c$.
For valid query and key positions, the masks are
\begin{equation}
\begin{aligned}
    M^{\mathrm{g}}_{c,s}
    &=
    \begin{cases}
        0, & s\leq c,\ D_s=D_c, \\
        -\infty, & \text{otherwise},
    \end{cases} \\
    M^{\mathrm{loc}}_{c,s}
    &=
    \begin{cases}
        0, & 0\leq c-s<w,\ D_s=D_c, \\
        -\infty, & \text{otherwise}.
    \end{cases}
\end{aligned}
\label{eq:gswa-training-masks}
\end{equation}
Thus, global attention accesses the observed prefix
within the same document, while local attention
accesses the current position and its preceding
$w-1$ positions.
We use $w=64$.
Queries and keys use the corresponding within-document
rotary positions consistently across loops.

For each later loop, a query-dependent sigmoid gate
combines the two outputs:
\begin{equation}
    \mathbf{A}_{\ell}^{(r)}
    =
    \bigl(1-\mathbf{G}_{\ell}^{(r)}\bigr)
        \odot\mathbf{A}_{\ell,\mathrm{g}}^{(r)}
    +
    \mathbf{G}_{\ell}^{(r)}
        \odot\mathbf{A}_{\ell,\mathrm{loc}}^{(r)}.
\label{eq:gswa-training-gate}
\end{equation}
The gate assigns one scalar to each sequence position
and attention head, broadcast across value features.
It is computed from normalized queries before rotary
position encoding, with gate parameters shared across
loops.
The first loop uses only the global branch.

\noindent\textbf{Token-parallel execution and gradients.}
Training first evaluates the first loop over the
entire teacher-forced sequence, retaining its
per-layer global KV tensors.
Subsequent loops are evaluated sequentially, with
all sequence positions processed in parallel
within each layer.
The attention masks enforce the logical cache
visibility directly on these sequence-level tensors.
The shared global KV tensors remain connected to
the computation graph, allowing gradients from
all subsequent uses to accumulate into the first-loop
representations.
The final-loop next-token loss is backpropagated
through both attention branches and all recurrent
and Foresight aggregation paths.

\noindent\textbf{Consistency with decoding.}
During decoding, each layer projects the active
loop inputs, adds the current first-loop KV to
the shared global bank, and adds each later loop's
KV to its local window before evaluating attention.
The current-cycle entries are therefore visible
under the same masks used during training.
This ordering requires layer-level coordination
while preserving concurrent execution across loops.
For identical input tokens, loop-ordered training
and cycle-ordered decoding evaluate the same
attention dependencies.
The bounded local caches apply to incremental
decoding; training retains the differentiable
activations required for backpropagation.